\documentclass[journal]{IEEEtai}

\usepackage[colorlinks,urlcolor=blue,linkcolor=blue,citecolor=blue]{hyperref}

\usepackage{color,array}
\usepackage{comment}
\usepackage{graphicx}
\usepackage{subcaption}
\usepackage{multirow}
\usepackage{xcolor}
\usepackage{acro}
\DeclareAcronym{ecblof}{short=ECBLOF, long=Enhanced Cluster-Based Local Outlier Factor}
\DeclareAcronym{lof}{short=LOF, long=Local Outlier Factor}
\DeclareAcronym{iforest}{short=IForest, long=Isolation Forest}
\DeclareAcronym{envelope}{short=EE, long=Elliptic Envelope}
\DeclareAcronym{autoencoder}{short=AE, long=AutoEncoder}
\DeclareAcronym{devnet}{short=DevNet, long=Deviation Network}
\DeclareAcronym{dagmm}{short=DAGMM, long=Deep Autoencoding Gaussian Mixture Model}
\DeclareAcronym{svm}{short=SVM, long=Support Vector Machine}
\DeclareAcronym{ocsvm}{short=OCSVM, long=One Class Support Vector Machine}
\DeclareAcronym{gan}{short=GAN, long=Generative Adversarial Network}
\DeclareAcronym{gnn}{short=GNN, long=Graph Neural Network}
\DeclareAcronym{ftt}{short=FTT, long=Feature Tokenizer + Transformer}
\DeclareAcronym{deepsad}{short=DeepSAD, long=Deep Semi-Supervised Anomaly Detection}
\DeclareAcronym{prenet}{short=PReNet, long=Progressive Recurrent Network}
\DeclareAcronym{mgbtai}{short=MGBTAI, long=Multi-Generations Binary Tree for Anomaly Identification}
\DeclareAcronym{dbtai}{short=d-BTAI, long=dynamic-Binary Tree Anomaly Identifier}
\DeclareAcronym{pef}{short=PEF, long=Parametric Elliot Function}
\DeclareAcronym{lstm}{short=LSTM, long=Long Short Term Memory}
\DeclareAcronym{qlstm}{short=q-LSTM, long=Quantile Long Short Term Memory}
\DeclareAcronym{qreg}{short=QReg, long=Deep Quantile Regression}
\DeclareAcronym{dl}{short=DL, long=Deep Learning}
\DeclareAcronym{ml}{short=ML, long=Machine Learning}

\title{Benchmarking Anomaly Detection Algorithms: Deep Learning and Beyond}

\begin{document}

\markboth{Journal of IEEE Transactions on Artificial Intelligence, Vol. 00, No. 0, Month 2020}
{First A. Author \MakeLowercase{\textit{et al.}}: Bare Demo of IEEEtai.cls for IEEE Journals of IEEE Transactions on Artificial Intelligence}

\author{
    Shanay Mehta$^{*}$, Shlok Mehendale$^{*}$, Nicole Fernandes$^{*\dagger}$, Jyotirmoy Sarkar$^{\ddagger}$,\\
    Santonu Sarkar and Snehanshu Saha%
    \thanks{$^{*}$Equal contribution.}
    \thanks{$^{\dagger}$Department of CS Don Bosco College of Engineering Goa, India (e-mail: nicoleana1204@gmail.com).}
    \thanks{$^{\ddagger}$GEHealthcare Bangalore, India (e-mail: jyotirmoy208@gmail.com).}
    \thanks{Department of Computer Science and Information Systems (CSIS), BITS Pilani K. K. Birla Goa Campus, Goa, India (e-mail: \{f20211322, f20221426, santonus, snehanshus\}@goa.bits-pilani.ac.in).}
}

\maketitle
\begin{abstract}
Detection of anomalous situations for complex mission-critical systems hold paramount importance when their service continuity needs to be ensured. A major challenge in detecting anomalies from the operational data arises due to the imbalanced class distribution problem since the anomalies are supposed to be rare events. This paper evaluates a diverse array of \ac{ml}-based anomaly detection algorithms through a comprehensive benchmark study. The paper contributes significantly by conducting an unbiased comparison of various anomaly detection algorithms, spanning classical \ac{ml}, including various tree-based approaches to \ac{dl} and outlier detection methods. The inclusion of 104 publicly available enhances the diversity of the study, allowing a more realistic evaluation of algorithm performance and emphasizing the importance of adaptability to real-world scenarios. 
    
The paper evaluates the general notion of \ac{dl} as a universal solution, showing that, while powerful, it is not always the best fit for every scenario. The findings reveal that recently proposed tree-based evolutionary algorithms match \ac{dl} methods and sometimes outperform them in many instances of univariate data where the size of the data is small and number of anomalies are less than $10\%$. Specifically, tree-based approaches successfully detect singleton anomalies in datasets where \ac{dl} falls short. 
%
To the best of the authors' knowledge, such a study on a large number of state-of-the-art algorithms using diverse data sets, with the objective of guiding researchers and practitioners in making informed algorithmic choices, has not been attempted earlier.
\end{abstract}

\begin{IEEEImpStatement}
Anomaly detection has gained momentum recently, backed by three decades of research. However, it has been observed that many state-of-the-art methods could surpass time-tested baselines only by a small margin. For an objective evaluation,  various anomaly detection techniques were compared on many datasets spanning diverse domains and data types. Through this diversity, the study identified their strengths and weaknesses and analyzed their generalizability. The study highlights the advantages of certain algorithms, ignored in the community, which train a model on a normal dataset without labeled anomalies. The study reveals that some recent SOTA methods rely on limiting assumptions, such as requiring at least two anomalies to be present in the data, requiring anomaly contamination (labels) in training data or a large fraction of data to be anomalies ($\sim$20\% in some cases). Finally, the study brings out a few methods, hitherto underscored compared to their more popular counterparts, which are far more generalist and diverse. The benchmarking study raises some important questions to be critically addressed in the context of anomaly detection such as the availability of well-annotated data, significant GPU resources and presence/absence of tiny percentage of anomalies.
\end{IEEEImpStatement}

\begin{IEEEkeywords}
Anomaly detection, Tree-based methods, Deep learning, Benchmarking 
\end{IEEEkeywords}
\section{Introduction}
Identification of anomalies or outliers has wide applicability, from fraud detection in financial transactions, security vulnerability identification, performance bottleneck analysis, industrial process monitoring, system reliability evaluation and so on. Nowadays, anomaly detection methods primarily use operational data of the system under consideration and construct a \ac{ml} model. 
A major challenge in building an effective ML model for anomaly detection is that the presence of anomalies in the dataset is rare. In streamlined industrial systems, a minority class instance (an anomalous condition) may occur once in thousands or millions of execution instances, causing a severe class imbalance for a classification-based ML model. 
This is a critical concern as, detecting anomalous conditions holds greater significance, making classification errors in this class more impactful than those in the majority class.

In recent years, deep neural networks have garnered significant attention in the anomaly detection field due to their impressive ability to automatically uncover complex data patterns and features. At the same time, tree-based ensemble methods such as Random Forests\cite{breiman2001random}, and Isolation Forests \cite{liu2008isolation} have been comparatively less discussed in the anomaly detection community. \textit{LinkedIn, for instance, still finds Isolation Forest based method to be quite effective in detecting various forms of abuse (modeled as anomalies) in their portal\cite{linkedin_anti_abuse}}.
Despite their efficacy in various domains, tree-based approaches have often operated in the shadow of \ac{dl} models, prompting us to revisit and reevaluate the efficacy of various \ac{ml} methods for anomaly detection.
This paper presents an unbiased and comprehensive comparison of various anomaly detection algorithms, encompassing a wide spectrum of methodologies. The summary of the findings is shown in Fig.~\ref{fig:summary}. 
Our contributions are as follows:
    \begin{figure*}[htbp]
        \centering
        \rotatebox{90}{ 
            \includegraphics[width=1.2\textwidth]{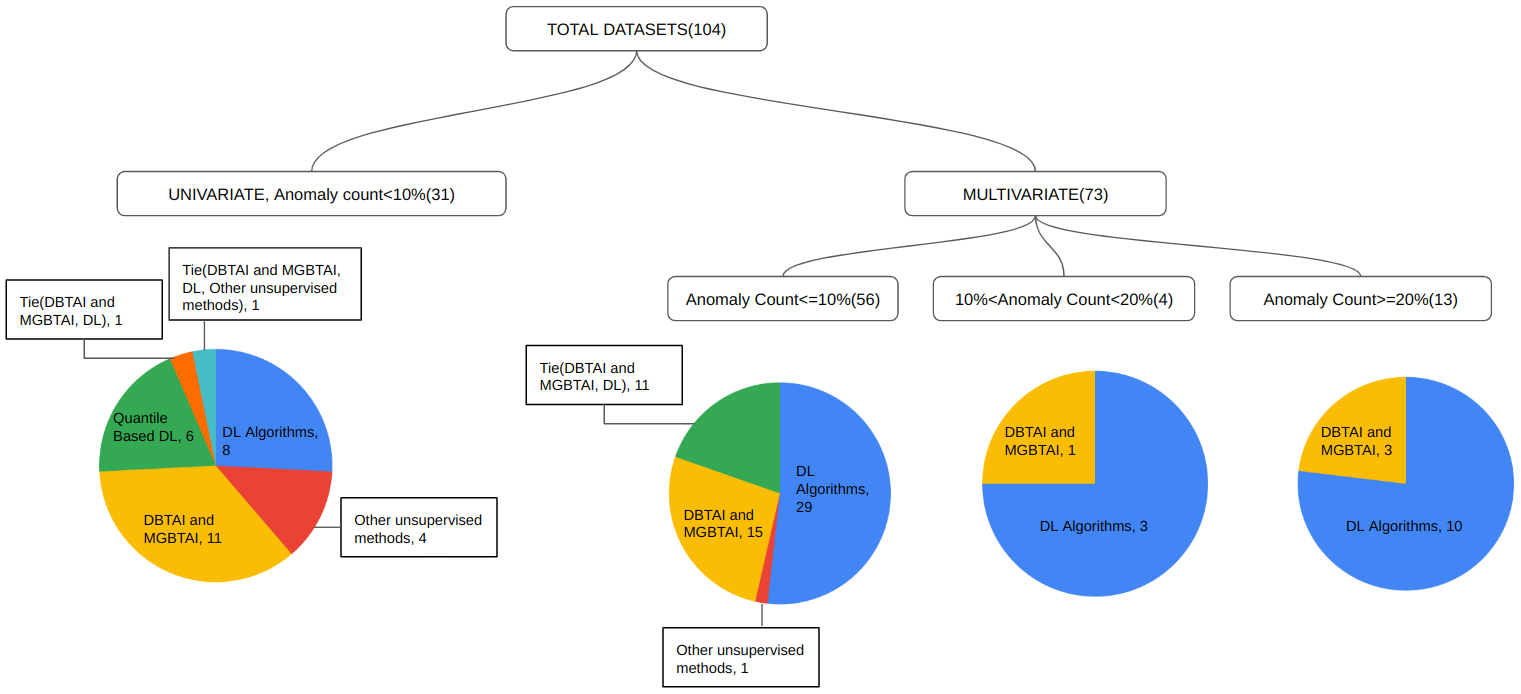}
        }
        \caption{\fontsize{9}{1em}\selectfont Comparison of anomaly detection algorithms. The algorithm obtaining the highest recall has been given credit. Note, DL algorithms in the figure consist of 11 recent SOTA methods.}
        \label{fig:summary}
    \end{figure*}
\begin{enumerate}
\item {\em Performance Evaluation}: The empirical study assesses a range of state-of-the-art (SOTA) \ac{ml} algorithms namely \ac{lof}\cite{breunig_lof_2000}, \ac{iforest}\cite{liu2008isolation}, \ac{ocsvm}\cite{heller_one_2003}, \ac{autoencoder}\cite{yin_anomaly_2022}, \ac{dagmm}\cite{zong_deep_2018}, \ac{lstm}\cite{hochreiter_long_1997}, \ac{qlstm}\cite{saha_quantile_2024}, \ac{qreg}\cite{tambuwal_deep_2021}, \ac{envelope}\cite{rousseeuw_fast_1999}, \ac{devnet}\cite{devnet}, \ac{gan}\cite{goodfellow2020generative}, \ac{gnn}\cite{deng_graph_2021}, Binary tree-based algorithms (\ac{mgbtai}\cite{sarkar_efficient_2023}, \ac{dbtai}\cite{sarkar_d-btai_2021}), and a few recent architectures like \ac{ftt}\cite{FTT}, \ac{deepsad}\cite{DeepSAD} and \ac{prenet}\cite{PReNet} for their efficacy in detecting anomalies from a large class of public datasets. The side-by-side comparison applied on datasets from different domains helps validate the algorithms' robustness and generalizability. 

\item {\em Resource Consumption}: Though \ac{dl} methods are known to be compute intensive, we documented in Section~\ref{sec:expt-rc} that algorithms like \ac{ftt}'s highly precise anomaly detection methods are a order of magnitude higher in training time, compared to any tree based methods. In contrast, we reported that tree based methods consume insignificant resources. 
This should force us to ponder about the effectiveness and suitability of \ac{dl} methods in resource constrained devices for anomaly detection. when an on-demand connectivity to a large computing infrastructure (such as cloud) is not an option.
\item {\em Generalizability of tree-based approach}: 
In this work, it was observed that tree-based evolutionary algorithms reported remarkable adaptability\cite{sarkar_efficient_2023}\cite{sarkar_d-btai_2021} where they excel in anomaly detection when there are rare as well as significant volumes of anomalies. The knee/elbow methods for these algorithms have been enhanced (Section~\ref{sec:expt-dbtai}) to determine thresholds which effectively identify the inflection point in anomaly score distributions. 
\item {\em Beyond \ac{dl}: The evaluation provides an alternate viewpoint- that \ac{dl} is not the panacea for anomaly detection. The findings 
demonstrate that while \ac{dl} is undoubtedly powerful, its performance may not consistently surpass that of other well-established methods.}
\item{\em Anomaly Prevalence and Detection Effectiveness}: The experiment demonstrated that the presence of a high percentage of anomalies (such as in the multivariate datasets used in this study) poses a unique challenge for anomaly detection, since an overwhelming abundance of anomalies can obscure the characteristics that distinguish them from normal data. 
It was empirically observed that most, well-known anomaly detection methods work well on datasets with less than 10\% anomalies. 
The study showed that with a smaller anomaly percentage, 
precision, recall, and the F1-score become more informative and reflective of a model's true anomaly detection capabilities. 
\end{enumerate}
\subsection*{Key Observations}In summary, the benchmarking study raises some important questions to be critically addressed in the context of anomaly detection such as (a) the availability of well-annotated data, (b) significant GPU resources and (c) presence/absence of tiny percentage of anomalies. For example, these recent SOTA \ac{dl} methods (FTTransformer, DeepSAD, PReNet) require take more than 3 hours to train a medium length dataset on Nvidia (Pascal) GPU P100 machine.

It is also to be noted that, these methods require labeled data for anomaly detection, which is not realistic in the context of the problem. 
It must not escape our attention that, an \textit{equitable} comparison between these recent \ac{dl} methods and tree based ones is not possible. This is because these recent \ac{dl} methods require a fraction of anomalies (labels) to be present in the training data for detection to be effective. On the contrary, the tree based methods don't require anomaly labels in the training data because they don't require training at all! It is interesting to observe that, slightly older SOTA \ac{dl} methods (Autoencoder, GAN, etc.) which do not require anomalies to be present in the training data fail to outperform tree based methods.

We must also note that all the three models (FTTransformer, PReNet, DeepSAD) which outperform the tree based method "need" anomalous contamination in training set i.e. these can't train on a singleton class. Hence, they do not work on datasets with a single anomaly. All univariate data sets have less than 20 anomalies and 5 among them have a single anomaly. Consequently, the performance of these aforementioned SOTA \ac{dl} methods in those cases drop significantly (for data sets: glass, lymphography, lympho\_h, wbc\_h, wine, WBC, WDBC, WPBC, smtp, hepatitis, ecoli, cmc etc.). 
An equitable comparison with the complex architectures (\ac{ftt}, \ac{deepsad} and \ac{prenet}) yields the following results on univariate data sets:  In Precision, tree based methods win on 19/29 datasets, tie on 4 and lose on 6, and in recall, tree based methods  win on 20/29 datasets, tie on 1 and lose on 8, to the aforementioned SOTA \ac{dl} methods (FTTransformer, PReNet, DeepSAD). Please see Section~\ref{sec:Experiment} details.
\section{Literature Review}
\label{sec:sec2}
Data-driven anomaly detection has been extensively researched for more than a decade. The report by 
~\cite{chandola_anomaly_2009} classified various types of anomalies and detection approaches before the prominence of \ac{ml} based approaches. Subsequently several academic surveys have been conducted on \ac{dl} based anomaly detection~\cite{chalapathy_deep_2019,pang_deep_2022}. 
There have been surveys on domain specific anomaly detection such as \ac{dl} based anomaly detection algorithms for images and videos~\cite{kiran_overview_2018,liu_deep_2023,cui_survey_2023}, network traffic~\cite{wang_machine_2021,eltanbouly_machine_2020}, real-time data~\cite{ariyaluran_habeeb_real-time_2019,cook_anomaly_2020,lu_review_2023}, urban traffice~\cite{djenouri_survey_2019} and so on.
In another recent survey, researchers have reviewed 52 algorithms and grouped them by statistics, density, distance, clustering, isolation, ensemble and so on~\cite{samariya_comprehensive_2023}. While this body of work provides a rich landscape of anomaly detection techniques, they do not deep-dive any of these methods to understand the efficacies and limitations through empirical studies.

Existing anomaly detection systems have used a combination of statistical methods~\cite{zhang_statistics-based_2012},
\ac{ml}algorithms\cite{ghosh_study_1999,bandaragoda_isolationbased_2018,castellani_real-world_2021,deng_graph_2021,erfani_high-dimensional_2016,luo_distributed_2018,falcao_quantitative_2019}, and domain-specific techniques. In particular, several different approaches including but not limited to, supervised methods, rule based classifiers, threshold and density based methods are prevalent. Each of these methods have their own hardships to overcome. Using industrial data, \cite{kotsiantis_supervised_2007} demonstrated that supervised anomaly detection faces challenges with scarce anomaly labels, adapting to new machine classes, and complex machinery. Rule-based classifier methods may struggle with unseen patterns, and ensemble techniques have limitations in memory and high-dimensional data. Anomaly detection systems that rely on static thresholds struggle to adapt to dynamic data containing  evolving patterns, resulting in high false positives and missed anomalies.  Density-based methods struggle with varying data densities, and profile-based techniques generate excessive false alarms in small datasets~\cite{karczmarek_k-means-based_2020}\cite{Wenli}. To address these issues, there is a growing need towards unsupervised and semi-supervised learning approaches. 
The work by~\cite{kiran_overview_2018} observed that unsupervised anomaly detection can be computationally intensive, especially in high-dimensional datasets. Clustering-based methods like K-Means \cite{kodinariya2013review} and isolation Forest \cite{liu2008isolation}  may require prior knowledge of cluster numbers. 
While Jacob and Tatbul \cite{jacob_exathlon_2021} investigated explainable anomaly detection in time series using real-world data, yet \ac{dl}-based time-series anomaly detection models were not explored well enough. This, significant growth in applying various ML algorithms to detect anomalies, and possibly the complexity and criticality of the anomaly detection problem triggered discussion on anomaly benchmarking data. In fact, there has been an avalanche of anomaly benchmarking data~\cite{han_songqiao_adbench_2022,paparrizos_tsb-uad_2022,wenig_timeeval_2022}, as well as empirical studies of the performances of existing algorithms \cite{
goix_how_2016,domingues_comparative_2018,huet_local_2022,schmidl_anomaly_2022} on different benchmark data. AD Bench \cite{han_songqiao_adbench_2022} examines the performance of 30 detection algorithms across an extensive set of 57 benchmark datasets, adding to the body of research in the field.
Wu and Keogh in~\cite{wu_current_2022}, introduced the UCR Time Series Anomaly
Archive, suggesting the inclusion of anomalies with a uniform random distribution in datasets for meaningful comparisons. 
 

In some literature, both shallow and deep-learning based anomaly detection methods are discussed, without any experimental results~\cite{DONG2021100379}. Researchers have assessed many unsupervised methods on small public dataset~\cite{campos_evaluation_2016,domingues_comparative_2018} and analyzed various characteristics such as scalability, memory consumption, and method robustness. Researchers have proposed a method to generate realistic synthetic anomaly data from real-world data~\cite{steinbuss_benchmarking_2021}. Paparrizos et al.~\cite{paparrizos_tsb-uad_2022} made a similar attempt to augment limited public datasets with synthetic ones. 

Researchers have investigated a suitable evaluation metric for ML methods in detecting anomalies. For instance, Doshi et al.~\cite{doshi_reward_2022} found surprising results, such as the widely used F1-score with point adjustment metric favouring a basic random guessing method over SOTA detectors. 
Kim et al.~\cite{kim_towards_2022} highlighted the flaws of the F1-score with point adjustment in both theoretical and experimental contexts. Although numerous new time series anomaly detection metrics have emerged, even the latest ones are not without their drawbacks. Another recent work reported that anomaly detection efficacy can be contaminated by modifying the train-test splitting~\cite{fourure_anomaly_2021}, if F1 and AVPR metrics are used. The work by Goix et al~\cite{goix_how_2016} proposes a new metric EM/MV to evaluate the performance of anomaly detection approaches. 

To conclude, with the availability of a substantial volume of anomaly benchmark data, and ML-driven anomaly detection methods in the public domain, recent benchmarking studies have focused \textit{only on deep-learning-based anomaly detection}. However, these studies did not analyze how these detection methods behave with the \textit{diversity (singleton, small, and significantly high numbers) in the anomalies and in the data} (type: univariate, multivariate, temporal, and non-temporal and size: (80 to 6,20,000 datapoints)). These studies did not revisit how anomalies should be identified when they are large in number. Moreover, these studies do not bring out the limitation of assuming anomalous contamination in training data, a debatable aspect in the \emph{dl} models. These observations lead to the following important questions.

What should be the percentage of anomalies in a dataset, beyond which they cease to become an outlier? 
The \ac{dl} community have demonstrated that anomaly detection using DL techniques works well for a certain class of data. In such a case, are we sure that non-DL algorithms are not performing better on those data sets?  Do we allow anomaly contamination i.e. labels while training? Does it not defeat the purpose of anomaly detection? While performance remains an importance metric, what should be the trade-off between performance and resource demand? The following sections investigates the performance of a few evolutionary, unsupervised approach (which do not require training, presence of minimum anomalies etc.) vis a vis a well-spread set of SOTA \ac{dl} based approaches on a reasonably large number of public data sets, from various application domains.

\section{Experiments and Results}
\label{sec:Experiment}

\begin{table*}[!hbt]
  \centering
  \begin{tabular}{|p{1.2cm}|p{16cm}|}
    \hline
    \textbf{Algorithm} & \textbf{Anomaly Detection Approach} \\
    \hline
    \ac{lof} & Uses data point densities to identify an anomaly by measuring how isolated a point is relative to its nearest neighbors in the feature space. Implemented using the Python Outlier Detection (PyOD)\cite{zhao_pyod_2019} library with default parameters. Trained on 70\% of data and tested on the entire dataset. \\
    \hline
    \ac{iforest} & Ensemble-based algorithm that isolates anomalies by constructing decision trees. Reported to perform well in high-dimensional data. The algorithm efficiently separates outliers by requiring fewer splits in the decision tree compared to normal data points. Implemented using scikit-learn\cite{pedregosa_scikit-learn_2011} with default parameters. Trained on 70\% of data and tested on the entire dataset. \\
    \hline
    \ac{ocsvm} & Constructs a hyperplane in a high-dimensional space to separate normal data from anomalies. Implemented using scikit-learn with default parameters. Trained on 70\% normal data, or whatever normal data was available. \\
    \hline
    \ac{autoencoder} & Anomalies are identified based on the reconstruction errors generated during the encoding-decoding process. \textit{Requires training on normal data.} Implemented using Keras\cite{chollet2015keras}. Lower threshold set at the 0.75th percentile, and upper threshold at the 99.25th percentile of the Mean Squared Error (MSE) values. Trained on 70\% normal data, or whatever normal data was available. Anomalies are detected by comparing the reconstruction error with the predefined thresholds. \\
    \hline
    \ac{dagmm} & Combines \ac{autoencoder} and Gaussian mixture models to model the data distribution and identify anomalies. It has a compression network to process low-dimensional representations, and the Gaussian mixture model helps capture data complexity. \textit{This algorithm requires at least 2 anomalies to be effective.} It is trained on 70\% normal data, or whatever normal data was available. Anomalies are detected by calculating anomaly scores, with thresholds set at two standard deviations above and below the mean anomaly score.\\
    \hline
    \ac{lstm} & Trained on a normal time-series data sequence. Acts as a predictor, and the prediction error, drawn from a multivariate Gaussian distribution, detects the likelihood of anomalous behavior. Implemented using Keras. Lower threshold set at the 5th percentile, and upper threshold at the 95th percentile of the Mean Squared Error (MSE) values. Trained on 70\% normal data, or whatever normal data was available. Anomalies are detected when the prediction error lies outside the defined thresholds. \\
    \hline
    \ac{qlstm} & Augments LSTM with quantile thresholds to define the range of normal behavior within the data. Implementation follows the methodology described in the authors' paper\cite{saha_quantile_2024}, which applies quantile thresholds to LSTM predictions. Anomalies are detected when the prediction error falls outside the defined quantile range. \\
    \hline
    \ac{qreg} & A multilayered LSTM-based RNN forecasts quantiles of the target distribution to detect anomalies. The core mathematical principle involves modeling the target variable's distribution using multiple quantile functions. Lower threshold set at the 0.9th percentile, and upper threshold at the 99.1st percentile of the predicted values. Anomalies are detected when the predicted value lies outside these quantile thresholds. Trained on 70\% data and tested on the entire dataset. \\
    \hline
    \ac{envelope} & Fits an ellipse around the central multivariate data points, isolating outliers. It needs a contamination parameter of 0.1 by default, with a support fraction of 0.75, and uses Mahalanobis distance for multivariate outlier detection. Implemented using default parameters from the sklearn package. Trained on 70\% of data and tested on the entire dataset. Anomalies are detected when data points fall outside the fitted ellipse. \\
    \hline
    \ac{devnet} & A \ac{dl}-based model designed specifically for anomaly detection tasks. Implemented using the \ac{dl}-based Outlier Detection (DeepOD)\cite{xu_deep_2023} library. Anomalies are detected based on the deviation score, with a threshold defined according to the model's performance and expected anomaly rate. Trained on 70\% data and tested on the entire dataset; \textit{it requires atleast 2 anomalies in its training set to function, and for optimal performance, it is recommended to include at least 2\% anomalies in the training data.} \\
    \hline
    \ac{gan} & Creates data distributions and detects anomalies by identifying data points that deviate from the generated distribution. It consists of generator and discriminator networks trained adversarially. Implemented using Keras. All data points whose discriminator score lies in the lowest 10th percentile are considered anomalies. Trained on 70\% normal data. \\
    \hline
    \ac{gnn} & GDN, which is based on graph neural networks, learns a graph of relationships between parameters and detects deviations from the patterns. Implementation follows the methodology described in the authors' paper\cite{deng_graph_2021}. \\
    \hline
    \ac{mgbtai} & An unsupervised approach that leverages a multi-generational binary tree structure to identify anomalies in data. Minimum clustering threshold set to 20\% of the dataset size and leaf level threshold set to 4. Used k-means clustering function. No training data required.\\
    \hline
    \ac{dbtai} & Like \ac{mgbtai}, it does not rely on training data. It adapts dynamically as data environments change.The small cluster threshold is set to 2\% of the data size. The leaf level threshold is set to 3. The minimum cluster threshold is set to 10\% of the data size and the number of clusters are 2 (for KMeans clustering at each split). The split threshold is 0.9 (used in the binary tree function). The anomaly threshold is determined dynamically using the knee/elbow method on the cumulative sum of sorted anomaly scores. The kernel density uses a gaussian kernel with default bandwidth and uses imbalance ratio to weight the density ratios. Used k-means clustering function. No training data required.\\
    \hline
    \ac{ftt} & \ac{ftt} is a sample adaptation of the original transformer architecture for tabular data. The model transforms all features (categorical and numerical) to embeddings and applies a stack of Transformer layers to the embeddings. However, as stated in the original paper's \cite{FTT} limitations: \ac{ftt} requires more resources (both hardware and time) for training than simple models such as ResNet and may not be easily scaled to datasets when the number of features is “too large”. \\
    \hline
    \ac{deepsad} & \ac{deepsad} is a generalization of the unsupervised Deep SVDD method to the semi-supervised anomaly detection setting and thus needs labeled data for training. It is also considered as an information-theoretic framework for deep anomaly detection.\\
    \hline
    \ac{prenet} & \ac{prenet} has a basic ResNet with input and output convolution layers, several residual blocks (ResBlocks) and a recurrent layer implemented using a LSTM. It is particularly created for the task of image deraining as mentioned in \cite{PReNet}.\\
    \hline
  \end{tabular}
  \caption{Descriptions and hyperparameter settings of SOTA algorithms benchmarked in this study}
  \label{tab:algo_list}
\end{table*}

\subsection{Dataset Descriptions}
A total of 73
multivariate datasets have been utilized in this study, sourced from \cite{han_songqiao_adbench_2022}, Singapore University of Technology and Design (SUTD) \cite{mathur_swat_2016} and the ODDS repository\footnote{\url{https://paperswithcode.com/dataset/odds}, last accessed on November 9, 2022.}. 6 of these 73 datasets are synthetically generated. Covering domains such as healthcare, finance, telecommunications, image analysis, and natural language processing, the compilation includes both time series and non-time series datasets, with seven datasets dedicated to temporal analysis. Data volumes range from 80 to 6,20,000 averaging 47,500 datapoints. The dimensions span from 3 to 1,555. Anomaly percentages range from 0.03\% to 43.51\%, with an average anomaly rate of 10\%.
Table 1 of the Supplementary provides details of the multivariate datasets including their size, dimensionality, number of anomalies, anomaly percentages, and corresponding domains.

The study also includes 31
univariate datasets (refer to
Table 10 of Supplementary). The collection includes both industrial as well as non-industrial datasets. Industrial datasets encompass the  Numenta Anomaly Benchmark(NAB)\footnote{\url{https://github.com/numenta/NAB/tree/master/data}} datasets and Yahoo Webscope\footnote{\url{https://webscope.sandbox.yahoo.com/}} datasets. Synthetic datasets, generated from AWS and Yahoo datasets, have also been included. The univariate datasets vary in size, ranging from 544 to 2,500, with an average of 1,688 datapoints. Anomaly percentages within these datasets range from 0.04\% to 1.47\%.

The study also ensures the inclusion of datasets from mission-critical systems. Storage data sets such as Yahoo! and AWS are extensively investigated. Additionally, Healthcare imaging device data which could afford no downtime and hence are mission-critical have also been investigated (Table 10 of the supplementary file, referred to as Industrial\_1 and Industrial\_2). Astronautics data which is also mission-critical was tested in this study.

A more detailed discussion of the characteristics of the datasets used in this study, along with the rationale and process behind generating synthetic data, can be found in Section 1.1 of the Supplementary.

\subsection{Anomaly Detection Algorithms}
The study selected a set of SOTA algorithms for anomaly detection to conduct a side-by-side assessment of their performances, focusing on highly cited algorithms and time-tested approaches. Details about all the algorithms benchmarked in this study, along with their hyperparameter settings, are given in Table~\ref{tab:algo_list}.

For transparency, all the codes and datasets are available in a completely anonymized repository\footnote{https://anonymous.4open.science/r/Anomaly-Benchmarking-776D/README.md}. Python version 3.10.12 was used. Compute-intensive algorithms, namely, \ac{devnet}, \ac{deepsad}, \ac{ftt} and \ac{prenet}  were run on Nvidia Tesla P100 GPU. All the other algorithms were run on CPU. The performance metrics used for algorithm evaluation are precision, recall, F1 score and AUC-ROC. 
\subsection{Resource Consumption} \label{sec:expt-rc}
\ac{dl} algorithms took significantly more time to run compared to classical and tree-based algorithms. Recent SOTA DL methods
(FTTransformer, DeepSAD, \ac{gan}, PReNet) require Nvidia (Pascal) GPU P100 (3584 Cores, 16GB RAM, 10.6 TeraFLOPS single
Precision) for training. Among them, \ac{ftt} was the most time-consuming, requiring around 30-40 minutes on average to train for each dataset. For larger datasets, the training time for DL algorithms such as GAN, DeepSAD, FTT, and PReNet exceeded 3 hours, at which point we terminated the training process. Consequently, these algorithms were not evaluated for those datasets, and their results were marked as NA. In contrast, classical and tree-based algorithms typically finished execution within 10-40 seconds, being trained on CPU devices with 8 GB RAM.


\subsection{Enhancing the \ac{dbtai} Algorithm}\label{sec:expt-dbtai}
Though \ac{dbtai} has shown promising results~\cite{sarkar_d-btai_2021} as an unsupervised anomaly detection method with sufficient generalizability, there are sufficient rooms for further improvement, which was carried out as a part of our study.
\subsubsection{Elimination of Manual Threshold}
\begin{figure}[h]
        \centering
        \includegraphics[width=0.35\textwidth]{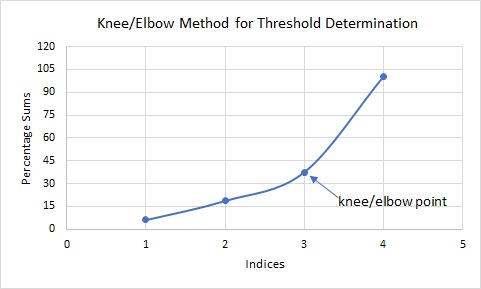} %
      \caption{Visualization of the elbow point}
      \label{knee_elbow}
\end{figure}
In the original approach, a datapoint $d_i\in D, i=1\cdots N$ with \ac{ecblof}\cite{sarkar_d-btai_2021} score $ec_i>$ a threshold $\psi$ is marked as an anomaly. This task is often tedious and non-adaptive, making the process inefficient for dynamic datasets. To address this problem an adaptive approach was adopted based on the knee/elbow method for determining the threshold. This approach identifies the knee/elbow point on a graph of sorted \ac{ecblof} scores. In the knee/elbow method, the idea is to find a point on the graph where the rate of change significantly slows down. This point is considered a potential threshold. To model this rate of change, a cumulative normalized sum of \ac{ecblof} values is computed as follows. 
\begin{enumerate}
    \item \ac{ecblof} values of all the datapoints are sorted $\{ec_i|i=1\cdots N\}$. 
    \item A cumulative \ac{ecblof} $EC_j$ for each datapoint $j$, is computed as $\sum_{i=1}^j ec_i$ and then the normalized cumulative \ac{ecblof} percentage sum is computed as 
    \[
    \overline{EC_j} = \frac{EC_j}{\max\{EC_1\cdots EC_N\}}\times 100
    \]
\end{enumerate}
The threshold computation is illustrated with an example. Consider four data points with \ac{ecblof} scores of 1, 2, 3, and 10. The cumulative $EC_i$ values are 1, 3, 6, and 16, and the normalized $\overline{EC_j}$ are 6.25\%, 18.75\%, 37.5\%, and 100\%. Plotting these values (as shown in Fig. \ref{knee_elbow}) where the X-axis is the $ec_i$ scores and the Y axis is the cumulative, normalized percentage scores $\overline{EC_i}$, the knee/elbow point is identified at (3, 37.5).
\subsubsection{Insufficiency of \ac{ecblof} Scores}

It may be noted that \ac{dbtai} uses clustering approach to detect anomalies. Even with adaptive threshold, \ac{ecblof} metric does not work well for large clusters. This is because large clusters naturally have lower scores since points tend to be closer to their centroids in dense regions, whereas small/sparse clusters have higher scores due to greater distances. This could lead to over-detection of anomalies in sparse regions and under-detection in dense regions. Weighting ECBLOF scores with density ratios helps address the imbalance in anomaly detection caused by varying cluster densities. By multiplying ECBLOF scores with density ratios, the approach scales the scores based on the relative density of each cluster. This normalization compensates for the inherent tendencies of the metric in different density regions. In dense clusters, the weighting increases the sensitivity to potential anomalies by amplifying scores, ensuring that true anomalies are not overlooked. In sparse clusters, it dampens the excessive scores, reducing false positives caused by natural sparsity. This mainly works very well for critically imbalanced datasets, as can be observed in the case of univariate datasets having very less number of anomalies. 

\subsection{Results - Multivariate Datasets}

In this subsection, the performance of anomaly detection algorithms is examined through Precision, Recall, F1-score, and AUC-ROC on multivariate datasets of varying sizes and domains. By assessing each of these metrics individually, the study provides insights into the strengths and weaknesses of the algorithms across different contexts, enabling a comprehensive evaluation of their performance.
The complete results for all algorithms on multivariate data can be found in Table 2, Table 3, Table 4 and Table 5 of the Supplementary.  


Fig.~\ref{fig:multivariate_comparison} illustrates the comparative performance of various anomaly detection algorithms across datasets, evaluated using four metrics: precision, recall, F1-score, and AUC-ROC. \textit{In case two or more algorithms achieve the highest value for a metric on a dataset, credit is assigned to each algorithm involved}.

 \ac{ftt} demonstrated the best precision, achieving the highest performance across 17 datasets, followed by \ac{mgbtai}, which excelled on 12 datasets. \ac{devnet} achieved the top precision on 11 datasets, while \ac{deepsad} and \ac{prenet} each performed well on 10 datasets. But, in general, \ac{devnet} is no longer more balanced than \ac{mgbtai}.

When examining recall, \ac{dbtai} performed the best, achieving the highest recall across 25 datasets. This is followed by \ac{deepsad} and \ac{devnet} , excelling on 21 and 20 datasets, respectively. Other notable performers include~\ac{ftt}, it obtained the highest recall on 12 datasets. \ac{mgbtai}, \ac{iforest} and \ac{prenet}, each excelled on 7 datasets.
\begin{figure*}[h!]
    \centering
    \includegraphics[width=0.8\linewidth]{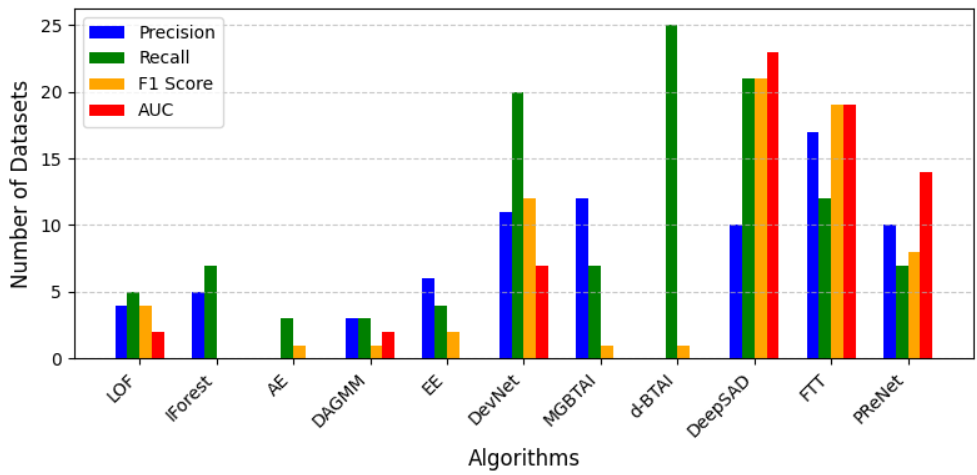} %
  \caption{Comparison of algorithms using various metrics on multivariate datasets}
  \label{fig:multivariate_comparison}
\end{figure*}
In terms of F1 Score, which balances both precision and recall, \ac{deepsad} remained dominant in 21 datasets, showcasing a strong overall ability to detect and classify anomalies. \ac{ftt} and \ac{devnet} also exhibited impressive performance, obtaining the highest F1 score on 19 and 12 datasets respectively. Conversely, other algorithms showed poor performance in F1-score indicating that they struggled to maintain both high precision and recall.

For AUC-ROC score, \ac{deepsad} emerged as the leading algorithm, excelling on 23 datasets, followed by \ac{ftt} at 19 datasets. \ac{prenet} also perfomred well achieving the highest AUC-ROC score on 14 datasets. On the lower end, some algorithms weren't able to obtain the highest AUC-ROC score on a single dataset, underlining their comparatively poor ability to discriminate. 

While algorithms like \ac{ftt}, \ac{deepsad}, and \ac{devnet} were strong performers across multiple metrics, \ac{mgbtai} and \ac{dbtai} distinguished themselves with their targeted excellence. \ac{mgbtai} performed impressively in precision, excelling on 12 datasets and showcasing its reliability in accurate anomaly detection. \ac{dbtai} stood out as a leader in recall, achieving the highest recall across 25 datasets, underlining its exceptional ability to detect anomalies comprehensively. 
Beyond these overall observations, we encountered some algorithm-specific behaviors that warranted closer examination. We observed distinctive performance characteristics in two algorithms, \ac{gan} and \ac{ocsvm} and we will now explore these scenarios.
\subsubsection{\textbf{Analysis of \ac{gan}}}
The study revealed that \ac{gan}s demonstrated some intriguing performance characteristics which needed to be discussed separately. A crucial aspect of \ac{gan}s' performance lies in its instability, as revealed through a comparative analysis of standard deviation values. The consistently high standard deviation[0-0.5774] 
(refer to Table 6 of Supplementary) on multiple runs highlights the algorithm's susceptibility to fluctuations, raising concerns about its stability and robustness in practical applications.
This instability is further emphasized by comparing \ac{gan} with \ac{autoencoder} which prove to be a more stable alternative. The standard deviation obtained from running \ac{gan} 3 times on all datasets went as high as 0.5774 whereas that obtained from running \ac{autoencoder} had a maximum value of 0.1985.

Furthermore, \ac{gan} achieved a perfect recall in 14 datasets. However, a closer examination revealed an interesting nuance: for 13 datasets, \ac{gan} classified all datapoints as anomalies. This peculiar outcome stemmed from the uniform discriminator score [1] 
assigned to all datapoints by the algorithm in the prediction of anomalies, emphasizing the need for a nuanced understanding of the algorithm's output.
One important observation is that \ac{gan} classified all datapoints as anomalies for all 6 SWaT datasets used in this study. SWaT datasets are multivariate time-series datasets. This raises a need to find a better algorithm for anomaly detection in multivariate time-series datasets.  \cite{deng_graph_2021} proposed a method based on \ac{gnn} to detect anomalies in multivariate time-series datasets. In our study, we have 7 multivariate time-series datasets, namely BATADAL\_04, SWaT 1 to SWaT 6. Additionally, 6 synthetic multivariate time series datasets were generated to further support the analysis of \ac{gnn}. The methodology for generating these datasets is detailed in Section 1.1 of the Supplementary. We observe that \ac{ftt} and \ac{mgbtai} outperform the other methods across these 13 datasets, and therefore, we present a comparative analysis of \ac{gnn}'s performance against \ac{ftt} and \ac{mgbtai}.

\ac{gnn} was not included in Fig.~\ref{fig:summary} and Fig.~\ref{fig:multivariate_comparison} because it is designed for multivariate time-series data only and hence is not tested on other datasets. As shown in Table 7 of the Supplementary, \ac{ftt} was the top-performing algorithm, achieving the best results on 6 datasets. It was followed by \ac{gnn}, which excelled on 4 datasets, and \ac{mgbtai}, which performed the best on 3 datasets. To determine the top performer, we first compared recall, and in the case of a tie, precision was used. Note that, since \ac{gan} classified all datapoints as anomalies, the recall was high and precision was remarkably low. This implies that \ac{gan} raises an alarming number of false positives therefore raising doubts about the  of the method itself. Therefore, in the overall comparison landscape (Fig.1), \ac{gan} was not included.
\subsubsection{\textbf{Analysis of \ac{ocsvm}}}
\ac{ocsvm} has proven to be a popular choice for anomaly detection due to its ability to model normal data and identify deviations effectively. However, an inherent challenge arises when the algorithm exhibits a high recall rate coupled with a low precision rate. Due to this behaviour, \ac{ocsvm} was analysed separately.
High recall implies that the \ac{ocsvm} is adept at capturing a significant portion of actual anomalies within the dataset. However, the drawback of this effectiveness lies in the concurrent low precision, indicating that a substantial number of normal instances are incorrectly classified as anomalies as seen in Table 8 and Table 9 of the Supplementary. The imbalance between recall and precision suggests that \ac{ocsvm} tends to be overly sensitive, marking a large proportion of instances as anomalies. 
While this sensitivity ensures that genuine anomalies are seldom overlooked, it compromises the precision of the model by introducing a considerable number of false positives.
\subsection{Results - Univariate datasets}
Apart from the multivariate datasets, the study includes 31 univariate datasets and tests 16 anomaly detection algorithms on these datasets. It also incorporates 4 quantile-based algorithms, 3 of these 4 algorithms are \ac{qlstm}~\cite{saha_quantile_2024} with sigmoid, tanh and \ac{pef} activation functions whereas the fourth quantile based algorithm is \ac{qreg}~\cite{tambuwal_deep_2021}. The performance of 16 anomaly detection algorithms on 31 univariate datasets is displayed in Table 11, Table 12, Table 13 and Table 14 of the Supplementary.
\begin{figure*}[t]
    \centering
    \includegraphics[width=0.8\linewidth]{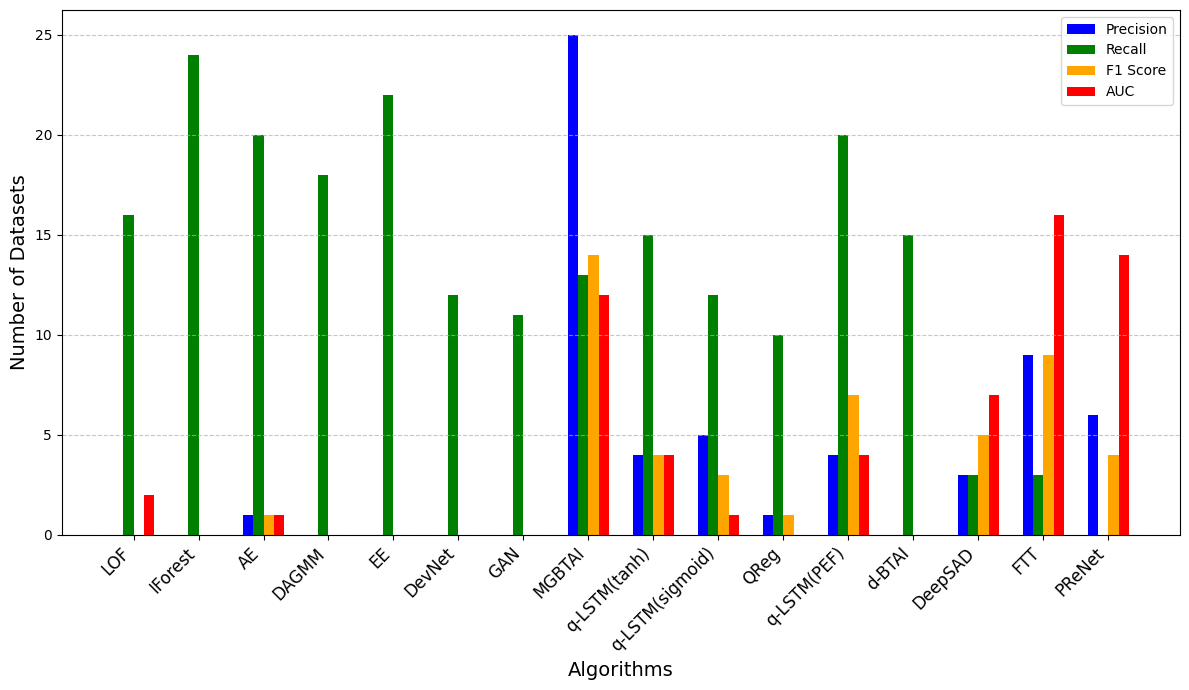} %
  \caption{Comparison of algorithms using various metrics on univariate datasets}
  \label{fig:univariate_comparison}
\end{figure*}
Fig.~\ref{fig:univariate_comparison} displays the performance of algorithms using the four metrics on univariate datasets.

For precision,  \ac{mgbtai} emerged as the best, achieving the highest precision in 25 datasets, followed by \ac{ftt}, which excelled in 9 datasets. \ac{prenet} also performed well in precision, in 6 datasets while the majority of the algorithms, showed poor precision.  Overall, \ac{mgbtai} leads in terms of precision.


In terms of recall, \ac{iforest} emerged as the top performer achieving the highest recall on 24 datasets. This was followed by \ac{envelope} with 22 datasets, and \ac{autoencoder} and \ac{qlstm}(\ac{pef}), both achieving top recall on 20 datasets. \ac{dagmm} secured the highest recall on 18 datasets respectively. \ac{qlstm}(tanh) and \ac{dbtai} achieved top recall on 15 datasets. \ac{devnet} and \ac{mgbtai} each performed well, achieving the highest recall on 13 datasets. We observed poor performance for \ac{ftt} and \ac{deepsad} peaking only in 2 datasets, whereas \ac{prenet} failed to achieve the highest recall on any dataset.

In terms of F1 score, \ac{mgbtai} emerged as the top performer, excelling on 12 datasets, followed by \ac{ftt}, which obtained the highest F1 Score on 9 datasets. Many traditional methods, including \ac{lof}, \ac{iforest}, \ac{dagmm}, Envelope, \ac{devnet}, and \ac{gan}, failed to make a impact.

In the AUC-ROC metric, \ac{ftt} and \ac{mgbtai} excelled, obtaining the highest score across 11 datasets. This is followed by \ac{prenet} achieving the highest AUC-ROC score on 7 datasets, highlighting their potential for robust anomaly detection. In contrast, many traditional methods, such as \ac{lof}, \ac{iforest}, and Envelope, had minimal or no impact, indicating limited effectiveness in achieving high discrimination accuracy.

The smaller size of univariate datasets presents unique challenges for data-hungry DL models, which rely on large amounts of data to learn complex patterns effectively. Consequently, in these scenarios, traditional unsupervised and tree-based approaches often achieve comparable or superior results.
\ac{mgbtai} stands out as a notable performer. While most algorithms excelled in recall, likely due to the small number of anomalies in the univariate datasets, the true challenge lies in achieving high precision. \ac{mgbtai} emerged as the best performer in this regard, achieving the highest precision on 25 datasets. What is particularly remarkable is that \ac{mgbtai}, a relatively lesser-known algorithm compared to other widely recognized techniques, consistently outperformed them. This impressive performance underscores \ac{mgbtai}'s potential as a highly effective anomaly detection algorithm, surpassing even some of the most popular methods in 'small-data' domain.

Complex architectures like \ac{ftt}, \ac{deepsad} and \ac{prenet} require a large amount of labeled data as well as some anomalous contamination in the traning set. This poses a major drawback as these SoTA deep learning algorithms cannot be used on datasets with critical anomaly percentages.
\subsection{Comparing Quantile Techniques}
The study evaluates the 4 quantile-based \ac{dl} algorithms: \ac{qreg}, \ac{qlstm} with tanh activation, \ac{qlstm} with sigmoid activation, and \ac{qlstm} with \ac{pef} activation. Table \ref{qlstm_perf_eval} presents the precision, recall, and AUC-ROC scores for \ac{qreg} and \ac{qlstm} with \ac{pef} on the univariate datasets. The full set of results, containing the precision, recall, F1 score, and AUC-ROC metrics for all 4 algorithms across the univariate datasets, is presented in Table 15 of the Supplementary Material. \ac{qlstm} with \ac{pef} emerged as the top performer for 15 datasets. \ac{qlstm} with tanh and sigmoid activation functions outperformed other algorithms on 5 datasets each whereas \ac{qreg} excelled on only 3 datasets.
\begin{table}[!htp]
\centering
\resizebox{0.5\textwidth}{!}{
\begin{tabular}{|c|ccc|ccc|}
\hline
\multirow{2}{*}{Dataset} & \multicolumn{3}{c|}{\ac{qreg}}                                  & \multicolumn{3}{c|}{\ac{qlstm}(\ac{pef})}                             \\ \cline{2-7} 
                         & \multicolumn{1}{c|}{Precision} & \multicolumn{1}{c|}{Recall} & AUC-ROC & \multicolumn{1}{c|}{Precision} & \multicolumn{1}{c|}{Recall} & AUC-ROC \\ \hline
yahoo2                    & \multicolumn{1}{c|}{0.1786}    & \multicolumn{1}{c|}{0.625}          & 0.8046  & \multicolumn{1}{c|}{1}               & \multicolumn{1}{c|}{0.375}         & 0.6875          \\ \hline
yahoo1                    & \multicolumn{1}{c|}{0}         & \multicolumn{1}{c|}{0}              & 0.4908  & \multicolumn{1}{c|}{0.0465}          & \multicolumn{1}{c|}{1}             & 0.9855          \\ \hline
yahoo3                    & \multicolumn{1}{c|}{0.1538}    & \multicolumn{1}{c|}{0.75}           & 0.8634  & \multicolumn{1}{c|}{\textbf{0.28}}   & \multicolumn{1}{c|}{1}             & 0.994           \\ \hline
yahoo5                    & \multicolumn{1}{c|}{0}         & \multicolumn{1}{c|}{0}              & 0.4865  & \multicolumn{1}{c|}{0.022}           & \multicolumn{1}{c|}{0.66}          & 0.7364          \\ \hline
yahoo6                    & \multicolumn{1}{c|}{0.0028}    & \multicolumn{1}{c|}{1}              & 0.5     & \multicolumn{1}{c|}{\textbf{0.0275}} & \multicolumn{1}{c|}{1}             & 0.95            \\ \hline
yahoo7                    & \multicolumn{1}{c|}{0}         & \multicolumn{1}{c|}{0}              & 0.4889  & \multicolumn{1}{c|}{0.066}           & \multicolumn{1}{c|}{0.54}          & 0.7448          \\ \hline
yahoo8                    & \multicolumn{1}{c|}{0}         & \multicolumn{1}{c|}{0}              & 0.4892  & \multicolumn{1}{c|}{0.028}           & \multicolumn{1}{c|}{0.3}           & 0.6188          \\ \hline
yahoo9                    & \multicolumn{1}{c|}{0.0049}    & \multicolumn{1}{c|}{\textbf{0.875}} & 0.5139  & \multicolumn{1}{c|}{0.0208}          & \multicolumn{1}{c|}{0.75}          & 0.7905          \\ \hline
Speed\_6005               & \multicolumn{1}{c|}{0}         & \multicolumn{1}{c|}{0}              & 0.4577  & \multicolumn{1}{c|}{\textbf{0.014}}  & \multicolumn{1}{c|}{1}             & 0.9718          \\ \hline
Speed\_7578               & \multicolumn{1}{c|}{0.5}       & \multicolumn{1}{c|}{0.24}           & 0.6121  & \multicolumn{1}{c|}{0.086}           & \multicolumn{1}{c|}{\textbf{1}}    & 0.981           \\ \hline
Speed\_t4013              & \multicolumn{1}{c|}{0.0392}    & \multicolumn{1}{c|}{1}              & 0.9902  & \multicolumn{1}{c|}{\textbf{0.053}}  & \multicolumn{1}{c|}{1}             & 0.9928          \\ \hline
TravelTime\_387           & \multicolumn{1}{c|}{0.0011}    & \multicolumn{1}{c|}{0.3333}         & 0.4782  & \multicolumn{1}{c|}{0.011}           & \multicolumn{1}{c|}{\textbf{0.67}} & 0.7988          \\ \hline
TravelTime\_451           & \multicolumn{1}{c|}{0}         & \multicolumn{1}{c|}{0}              & 0.4903  & \multicolumn{1}{c|}{0.006}           & \multicolumn{1}{c|}{1}             & 0.9616          \\ \hline
Occupancy\_6005           & \multicolumn{1}{c|}{0}         & \multicolumn{1}{c|}{0}              & 0.4891  & \multicolumn{1}{c|}{\textbf{0.03}}   & \multicolumn{1}{c|}{1}             & 0.9932          \\ \hline
Occupancy\_t4013          & \multicolumn{1}{c|}{0.0038}    & \multicolumn{1}{c|}{1}              & 0.895   & \multicolumn{1}{c|}{0.06}            & \multicolumn{1}{c|}{1}             & 0.9937          \\ \hline
yahoo\_syn1               & \multicolumn{1}{c|}{0.0303}    & \multicolumn{1}{c|}{0.0833}         & 0.5303  & \multicolumn{1}{c|}{\textbf{0.375}}  & \multicolumn{1}{c|}{1}             & 0.9928          \\ \hline
yahoo\_syn2               & \multicolumn{1}{c|}{0}         & \multicolumn{1}{c|}{0}              & 0.4903  & \multicolumn{1}{c|}{\textbf{1}}      & \multicolumn{1}{c|}{0.611}         & 0.8055          \\ \hline
yahoo\_syn3               & \multicolumn{1}{c|}{0.0805}    & \multicolumn{1}{c|}{0.3889}         & 0.6664  & \multicolumn{1}{c|}{0.6}             & \multicolumn{1}{c|}{\textbf{1}}    & 0.9958          \\ \hline
yahoo\_syn5               & \multicolumn{1}{c|}{0}         & \multicolumn{1}{c|}{0}              & 0.4908  & \multicolumn{1}{c|}{0.0625}          & \multicolumn{1}{c|}{0.578}         & 0.7306          \\ \hline
yahoo\_syn6               & \multicolumn{1}{c|}{0.0098}    & \multicolumn{1}{c|}{\textbf{1}}     & 0.5     & \multicolumn{1}{c|}{0.764}           & \multicolumn{1}{c|}{0.928}         & 0.9625          \\ \hline
yahoo\_syn7               & \multicolumn{1}{c|}{0.0092}    & \multicolumn{1}{c|}{0.1905}         & 0.4662  & \multicolumn{1}{c|}{0.411}           & \multicolumn{1}{c|}{\textbf{0.66}} & 0.824           \\ \hline
yahoo\_syn8               & \multicolumn{1}{c|}{0.2}       & \multicolumn{1}{c|}{\textbf{1}}     & 0.65    & \multicolumn{1}{c|}{0.197}           & \multicolumn{1}{c|}{0.7}           & 0.8329          \\ \hline
yahoo\_syn9               & \multicolumn{1}{c|}{0.0102}    & \multicolumn{1}{c|}{0.8889}         & 0.4789  & \multicolumn{1}{c|}{1}               & \multicolumn{1}{c|}{\textbf{0.94}} & 0.97            \\ \hline
aws1                      & \multicolumn{1}{c|}{0.0204}    & \multicolumn{1}{c|}{1}              & 0.977   & \multicolumn{1}{c|}{\textbf{0.041}}  & \multicolumn{1}{c|}{1}             & 0.9888          \\ \hline
aws2                      & \multicolumn{1}{c|}{0}         & \multicolumn{1}{c|}{0}              & 0.4895  & \multicolumn{1}{c|}{0.0042}          & \multicolumn{1}{c|}{1}             & 0.9045          \\ \hline
aws3                      & \multicolumn{1}{c|}{0.0116}    & \multicolumn{1}{c|}{1}              & 0.9716  & \multicolumn{1}{c|}{\textbf{0.0181}} & \multicolumn{1}{c|}{1}             & 0.9818          \\ \hline
aws\_syn1                 & \multicolumn{1}{c|}{0.0204}    & \multicolumn{1}{c|}{1}              & 0.977   & \multicolumn{1}{c|}{0.224}           & \multicolumn{1}{c|}{1}             & 0.9833          \\ \hline
aws\_syn2                 & \multicolumn{1}{c|}{0}         & \multicolumn{1}{c|}{0}              & 0.4895  & \multicolumn{1}{c|}{0.0489}          & \multicolumn{1}{c|}{1}             & 0.9211          \\ \hline
aws\_syn3                 & \multicolumn{1}{c|}{0.0116}    & \multicolumn{1}{c|}{1}              & 0.9716  & \multicolumn{1}{c|}{1}               & \multicolumn{1}{c|}{\textbf{1}}    & 1               \\ \hline
Industrial\_1             & \multicolumn{1}{c|}{0}         & \multicolumn{1}{c|}{0}              & 0.48    & \multicolumn{1}{c|}{0.357}           & \multicolumn{1}{c|}{\textbf{1}}    & 0.9865          \\ \hline
Industrial\_2             & \multicolumn{1}{c|}{0.0148}    & \multicolumn{1}{c|}{1}              & 0.9378  & \multicolumn{1}{c|}{1}               & \multicolumn{1}{c|}{\textbf{1}}    & 1               \\ \hline
\end{tabular}%
}
\caption{Performance evaluation of \ac{qreg} and \ac{qlstm} with \ac{pef} activation function on 31 univariate datasets. The best performing algorithm(s) is marked in bold.}
\label{qlstm_perf_eval}
\end{table}

\subsection{Experimentation with \ac{pef}}
\ac{pef} is a modification of the conventional activation function used in \ac{lstm} models. It has a slower saturation rate compared to other activation functions like sigmoid and tanh, which helps in overcoming the saturation problem during training and prediction. \textit{The \ac{pef} activation function was integrated into \ac{lstm}, \ac{qreg}, \ac{gan} and \ac{autoencoder}}, comparing their performance with models using standard tanh and sigmoid activation functions across 31 univariate datasets.
This comparison has been made with respect to recall, in case of a tie between the algorithms precision is taken into consideration and in case of further tie AUC-ROC values are considered.
\\
\subsubsection{\ac{lstm}}
\ac{lstm} with standard tanh and sigmoid activations \textit{performed better} in 11 datasets, while that with \ac{pef} excelled in 10 datasets as seen in Table \ref{lstm_pef}. \textit{Both the algorithms performed equally well on the remaining 10 datasets.}
\begin{table}[!htp]
\centering
\resizebox{0.5\textwidth}{!}{
\begin{tabular}{|c|ccc|ccc|}
\hline
\multirow{2}{*}{Dataset} & \multicolumn{3}{c|}{\ac{lstm}}                                  & \multicolumn{3}{c|}{\ac{lstm}(\ac{pef})}                             \\ \cline{2-7} 
                         & \multicolumn{1}{c|}{Precision} & \multicolumn{1}{c|}{Recall} & AUC-ROC & \multicolumn{1}{c|}{Precision} & \multicolumn{1}{c|}{Recall} & AUC-ROC \\ \hline
yahoo1                            & \multicolumn{1}{c|}{0.007}              & \multicolumn{1}{c|}{0.5}                   & 0.7002           & \multicolumn{1}{c|}{0.007}              & \multicolumn{1}{c|}{0.5}                   & 0.7002           \\ \hline
yahoo2                            & \multicolumn{1}{c|}{0.0411}             & \multicolumn{1}{c|}{0.75}                  & 0.8267           & \multicolumn{1}{c|}{0.0411}             & \multicolumn{1}{c|}{0.75}                  & 0.8267           \\ \hline
yahoo3                            & \multicolumn{1}{c|}{0.0417}             & \multicolumn{1}{c|}{0.75}                  & 0.8267           & \multicolumn{1}{c|}{0.0486}             & \multicolumn{1}{c|}{\textbf{0.875}}        & 0.8895           \\ \hline
yahoo5                            & \multicolumn{1}{c|}{0.0141}             & \multicolumn{1}{c|}{0.2222}                & 0.5614           & \multicolumn{1}{c|}{0.0352}             & \multicolumn{1}{c|}{\textbf{0.5556}}       & 0.7292           \\ \hline
yahoo6                            & \multicolumn{1}{c|}{0.0141}             & \multicolumn{1}{c|}{0.5}                   & 0.7005           & \multicolumn{1}{c|}{0.0211}             & \multicolumn{1}{c|}{\textbf{0.75}}         & 0.8258           \\ \hline
yahoo7                            & \multicolumn{1}{c|}{0.0119}             & \multicolumn{1}{c|}{\textbf{0.1818}}       & 0.5411           & \multicolumn{1}{c|}{0.006}              & \multicolumn{1}{c|}{0.0909}                & 0.4953           \\ \hline
yahoo8                            & \multicolumn{1}{c|}{0.0119}             & \multicolumn{1}{c|}{\textbf{0.2}}          & 0.5502           & \multicolumn{1}{c|}{0.006}              & \multicolumn{1}{c|}{0.1}                   & 0.4999           \\ \hline
yahoo9                            & \multicolumn{1}{c|}{0.0119}             & \multicolumn{1}{c|}{\textbf{0.25}}         & 0.5753           & \multicolumn{1}{c|}{0}                  & \multicolumn{1}{c|}{0}                     & 0.4497           \\ \hline
Speed\_6005                       & \multicolumn{1}{c|}{0.004}              & \multicolumn{1}{c|}{\textbf{1}}            & 0.9497           & \multicolumn{1}{c|}{0}                  & \multicolumn{1}{c|}{0}                     & 0.4499           \\ \hline
Speed\_7578                       & \multicolumn{1}{c|}{0.0254}             & \multicolumn{1}{c|}{0.75}                  & 0.8237           & \multicolumn{1}{c|}{\textbf{0.0263}}    & \multicolumn{1}{c|}{0.75}                  & 0.8254           \\ \hline
Speed\_t4013                      & \multicolumn{1}{c|}{\textbf{0.0079}}    & \multicolumn{1}{c|}{1}                     & 0.9494           & \multicolumn{1}{c|}{0.0078}             & \multicolumn{1}{c|}{1}                     & 0.9492           \\ \hline
TravelTime\_387                   & \multicolumn{1}{c|}{0}                  & \multicolumn{1}{c|}{0}                     & 0.4499           & \multicolumn{1}{c|}{0.004}              & \multicolumn{1}{c|}{\textbf{0.3333}}       & 0.6167           \\ \hline
TravelTime\_451                   & \multicolumn{1}{c|}{0}                  & \multicolumn{1}{c|}{0}                     & 0.45             & \multicolumn{1}{c|}{0}                  & \multicolumn{1}{c|}{0}                     & 0.45             \\ \hline
Occupancy\_6005                   & \multicolumn{1}{c|}{0.0042}             & \multicolumn{1}{c|}{1}                     & 0.9501           & \multicolumn{1}{c|}{0.0042}             & \multicolumn{1}{c|}{1}                     & 0.9501           \\ \hline
Occupancy\_t4013                  & \multicolumn{1}{c|}{0.008}              & \multicolumn{1}{c|}{1}                     & 0.9503           & \multicolumn{1}{c|}{0.008}              & \multicolumn{1}{c|}{1}                     & 0.9503           \\ \hline
yahoo\_syn1                       & \multicolumn{1}{c|}{0.0211}             & \multicolumn{1}{c|}{0.25}                  & 0.5755           & \multicolumn{1}{c|}{0.0282}             & \multicolumn{1}{c|}{\textbf{0.3333}}       & 0.6176           \\ \hline
yahoo\_syn2                       & \multicolumn{1}{c|}{0.0411}             & \multicolumn{1}{c|}{0.3333}                & 0.6181           & \multicolumn{1}{c|}{0.0479}             & \multicolumn{1}{c|}{\textbf{0.3889}}       & 0.6462           \\ \hline
yahoo\_syn3                       & \multicolumn{1}{c|}{0.0479}             & \multicolumn{1}{c|}{0.3889}                & 0.6458           & \multicolumn{1}{c|}{0.0548}             & \multicolumn{1}{c|}{\textbf{0.4444}}       & 0.6739           \\ \hline
yahoo\_syn5                       & \multicolumn{1}{c|}{0.0208}             & \multicolumn{1}{c|}{0.1579}                & 0.5289           & \multicolumn{1}{c|}{0.0278}             & \multicolumn{1}{c|}{\textbf{0.2105}}       & 0.5556           \\ \hline
yahoo\_syn6                       & \multicolumn{1}{c|}{0.0278}             & \multicolumn{1}{c|}{0.2857}                & 0.5934           & \multicolumn{1}{c|}{0.0278}             & \multicolumn{1}{c|}{0.2857}                & 0.5934           \\ \hline
yahoo\_syn7                       & \multicolumn{1}{c|}{0.0118}             & \multicolumn{1}{c|}{0.0118}                & 0.4972           & \multicolumn{1}{c|}{0.0059}             & \multicolumn{1}{c|} {\textbf{0.0476}}      & 0.4731           \\ \hline
yahoo\_syn8                       & \multicolumn{1}{c|}{0.0118}             & \multicolumn{1}{c|}{\textbf{0.1}}          & 0.4996           & \multicolumn{1}{c|}{0.0059}             & \multicolumn{1}{c|}{0.05}                  & 0.4743           \\ \hline
yahoo\_syn9                       & \multicolumn{1}{c|}{0.0176}             & \multicolumn{1}{c|}{\textbf{0.1667}}       & 0.5333           & \multicolumn{1}{c|}{0.0059}             & \multicolumn{1}{c|}{0.0556}                & 0.4771           \\ \hline
aws1                              & \multicolumn{1}{c|}{0.0094}             & \multicolumn{1}{c|}{1}                     & 0.9498           & \multicolumn{1}{c|}{0.0094}             & \multicolumn{1}{c|}{1}                     & 0.9498           \\ \hline
aws2                              & \multicolumn{1}{c|}{0.0032}             & \multicolumn{1}{c|}{0.5}                   & 0.6865           & \multicolumn{1}{c|}{0.0032}             & \multicolumn{1}{c|}{0.5}                   & 0.6978           \\ \hline
aws3                              & \multicolumn{1}{c|}{0.0067}             & \multicolumn{1}{c|}{\textbf{1}}            & 0.9502           & \multicolumn{1}{c|}{0}                  & \multicolumn{1}{c|}{0}                     & 0.4498           \\ \hline
aws\_syn1                         & \multicolumn{1}{c|}{0.0935}             & \multicolumn{1}{c|}{1}                     & 0.9533           & \multicolumn{1}{c|}{0.0935}             & \multicolumn{1}{c|}{1}                     & 0.9533           \\ \hline
aws\_syn2                         & \multicolumn{1}{c|}{\textbf{0.0249}}    & \multicolumn{1}{c|}{1}                     & 0.8412           & \multicolumn{1}{c|}{0.0176}             & \multicolumn{1}{c|}{1}                     & 0.7733           \\ \hline
aws\_syn3                         & \multicolumn{1}{c|}{0.0667}             & \multicolumn{1}{c|}{1}                     & 0.953            & \multicolumn{1}{c|}{0.0667}             & \multicolumn{1}{c|}{1}                     & 0.953            \\ \hline
Industrial\_1                     & \multicolumn{1}{c|}{0.014}              & \multicolumn{1}{c|}{\textbf{1}}            & 0.9343           & \multicolumn{1}{c|}{0}                  & \multicolumn{1}{c|}{0}                     & 0.42             \\ \hline
Industrial\_2                     & \multicolumn{1}{c|}{0.0976}             & \multicolumn{1}{c|}{\textbf{1}}            & 0.931            & \multicolumn{1}{c|}{0.11}               & \multicolumn{1}{c|}{0.75}                  & 0.83  \\ \hline         
\end{tabular}
}
\caption{A comparative study of \ac{lstm} versus \ac{lstm} using \ac{pef} activation function on 31 univariate datasets. The best performing algorithm(s) is marked in bold.}
\label{lstm_pef}
\end{table}
\\
\subsubsection{\ac{qreg}}
As shown in Table \ref{dqr_pef}, \ac{qreg} \textit{using standard activation functions excelled} in 17 datasets whereas \ac{qreg} with \ac{pef} performed well on 14 datasets.
\begin{table}[!htp]
 \centering
 \resizebox{0.5\textwidth}{!}{
\begin{tabular}{|c|ccc|ccc|}
\hline
\multirow{2}{*}{Dataset} & \multicolumn{3}{c|}{\ac{qreg}}                                                                                    & \multicolumn{3}{c|}{\ac{qreg}(\ac{pef})}                                                                                  \\ \cline{2-7} 
                                  & \multicolumn{1}{c|}{Precision} & \multicolumn{1}{c|}{Recall} & AUC-ROC & \multicolumn{1}{c|}{Precision} & \multicolumn{1}{c|}{Recall} & AUC-ROC \\ \hline
yahoo1                            & \multicolumn{1}{c|}{0}                  & \multicolumn{1}{c|}{0}                     & 0.4908           & \multicolumn{1}{c|}{0.0014}             & \multicolumn{1}{c|}{\textbf{1}}            & 0.5              \\ \hline
yahoo2                            & \multicolumn{1}{c|}{0.1786}             & \multicolumn{1}{c|}{0.625}                 & 0.8046           & \multicolumn{1}{c|}{0.0055}             & \multicolumn{1}{c|}{\textbf{1}}            & 0.5              \\ \hline
yahoo3                            & \multicolumn{1}{c|}{0.1538}             & \multicolumn{1}{c|}{\textbf{0.75}}         & 0.8634           & \multicolumn{1}{c|}{0.0169}             & \multicolumn{1}{c|}{0.125}                 & 0.5422           \\ \hline
yahoo5                            & \multicolumn{1}{c|}{0}                  & \multicolumn{1}{c|}{0}                     & 0.4865           & \multicolumn{1}{c|}{0.0143}             & \multicolumn{1}{c|}{\textbf{0.444}}        & 0.6246           \\ \hline
yahoo6                            & \multicolumn{1}{c|}{0.0028}             & \multicolumn{1}{c|}{\textbf{1}}            & 0.5              & \multicolumn{1}{c|}{0.0032}             & \multicolumn{1}{c|}{0.75}                  & 0.5423           \\ \hline
yahoo7                            & \multicolumn{1}{c|}{0}                  & \multicolumn{1}{c|}{0}                     & 0.4889           & \multicolumn{1}{c|}{0.0104}             & \multicolumn{1}{c|}{\textbf{0.1818}}       & 0.5336           \\ \hline
yahoo8                            & \multicolumn{1}{c|}{0}                  & \multicolumn{1}{c|}{0}                     & 0.4892           & \multicolumn{1}{c|}{0.008}              & \multicolumn{1}{c|}{\textbf{0.8}}          & 0.6022           \\ \hline
yahoo9                            & \multicolumn{1}{c|}{0.0049}             & \multicolumn{1}{c|}{\textbf{0.875}}        & 0.5139           & \multicolumn{1}{c|}{0}                  & \multicolumn{1}{c|}{0}                     & 0.4892           \\ \hline
Speed\_6005                       & \multicolumn{1}{c|}{0}                  & \multicolumn{1}{c|}{0}                     & \textbf{0.4577}  & \multicolumn{1}{c|}{0}                  & \multicolumn{1}{c|}{0}                     & 0.3684           \\ \hline
Speed\_7578                       & \multicolumn{1}{c|}{\textbf{0.5}}       & \multicolumn{1}{c|}{0.25}                  & 0.6121           & \multicolumn{1}{c|}{0.0022}             & \multicolumn{1}{c|}{0.25}                  & 0.4241           \\ \hline
Speed\_t4013                      & \multicolumn{1}{c|}{0.0392}             & \multicolumn{1}{c|}{\textbf{1}}            & 0.9902           & \multicolumn{1}{c|}{0.0009}             & \multicolumn{1}{c|}{0.5}                   & 0.5275           \\ \hline
TravelTime\_387                   & \multicolumn{1}{c|}{0.0011}             & \multicolumn{1}{c|}{0.3333}                & 0.4782           & \multicolumn{1}{c|}{\textbf{0.0037}}    & \multicolumn{1}{c|}{0.3333}                & 0.6123           \\ \hline
TravelTime\_451                   & \multicolumn{1}{c|}{0}                  & \multicolumn{1}{c|}{0}                     & \textbf{0.4903}  & \multicolumn{1}{c|}{0}                  & \multicolumn{1}{c|}{0}                     & 0.4541           \\ \hline
Occupancy\_6005                   & \multicolumn{1}{c|}{0}                  & \multicolumn{1}{c|}{0}                     & 0.4891           & \multicolumn{1}{c|}{0}                  & \multicolumn{1}{c|}{0}                     & \textbf{0.4901}  \\ \hline
Occupancy\_t4013                  & \multicolumn{1}{c|}{0.0038}             & \multicolumn{1}{c|}{\textbf{1}}            & 0.895            & \multicolumn{1}{c|}{0.0208}             & \multicolumn{1}{c|}{0.5}                   & 0.7406           \\ \hline
yahoo\_syn1                       & \multicolumn{1}{c|}{0.0303}             & \multicolumn{1}{c|}{0.0833}                & 0.5303           & \multicolumn{1}{c|}{0.0085}             & \multicolumn{1}{c|}{\textbf{1}}            & 0.5              \\ \hline
yahoo\_syn2                       & \multicolumn{1}{c|}{0}                  & \multicolumn{1}{c|}{0}                     & 0.4903           & \multicolumn{1}{c|}{0.0123}             & \multicolumn{1}{c|}{\textbf{1}}            & 0.5              \\ \hline
yahoo\_syn3                       & \multicolumn{1}{c|}{0.0805}             & \multicolumn{1}{c|}{\textbf{0.3889}}       & 0.664            & \multicolumn{1}{c|}{0.0521}             & \multicolumn{1}{c|}{0.2778}                & 0.607            \\ \hline
yahoo\_syn5                       & \multicolumn{1}{c|}{0}                  & \multicolumn{1}{c|}{0}                     & 0.4908           & \multicolumn{1}{c|}{0.0269}             & \multicolumn{1}{c|}{\textbf{0.2632}}       & 0.5673           \\ \hline
yahoo\_syn6                       & \multicolumn{1}{c|}{0.0098}             & \multicolumn{1}{c|}{\textbf{1}}            & 0.5              & \multicolumn{1}{c|}{0.0117}             & \multicolumn{1}{c|}{0.2143}                & 0.5177           \\ \hline
yahoo\_syn7                       & \multicolumn{1}{c|}{0.0092}             & \multicolumn{1}{c|}{\textbf{0.1905}}       & 0.4662           & \multicolumn{1}{c|}{0.023}              & \multicolumn{1}{c|}{0.0952}                & 0.5221           \\ \hline
yahoo\_syn8                       & \multicolumn{1}{c|}{0.2}                & \multicolumn{1}{c|}{\textbf{1}}            & 0.65             & \multicolumn{1}{c|}{0.0147}             & \multicolumn{1}{c|}{0.05}                  & 0.5049           \\ \hline
yahoo\_syn9                       & \multicolumn{1}{c|}{0.0102}             & \multicolumn{1}{c|}{\textbf{0.8889}}       & 0.4789           & \multicolumn{1}{c|}{0}                  & \multicolumn{1}{c|}{0}                     & 0.4901           \\ \hline
aws1                              & \multicolumn{1}{c|}{0.0204}             & \multicolumn{1}{c|}{\textbf{1}}            & 0.977            & \multicolumn{1}{c|}{0}                  & \multicolumn{1}{c|}{0}                     & 0.4167           \\ \hline
aws2                              & \multicolumn{1}{c|}{0}                  & \multicolumn{1}{c|}{0}                     & 0.4895           & \multicolumn{1}{c|}{0.0008}             & \multicolumn{1}{c|}{\textbf{1}}            & 0.5              \\ \hline
aws3                              & \multicolumn{1}{c|}{\textbf{0.0116}}    & \multicolumn{1}{c|}{1}                     & 0.9716           & \multicolumn{1}{c|}{0.0027}             & \multicolumn{1}{c|}{1}                     & 0.8776           \\ \hline
aws\_syn1                         & \multicolumn{1}{c|}{0.0204}             & \multicolumn{1}{c|}{0.1}                   & 0.5268           & \multicolumn{1}{c|}{\textbf{0.0238}}    & \multicolumn{1}{c|}{0.1}                   & 0.5302           \\ \hline
aws\_syn2                         & \multicolumn{1}{c|}{0.0217}             & \multicolumn{1}{c|}{0.05}                  & 0.5159           & \multicolumn{1}{c|}{0.0081}             & \multicolumn{1}{c|}{\textbf{1}}            & 0.5              \\ \hline
aws\_syn3                         & \multicolumn{1}{c|}{0}                  & \multicolumn{1}{c|}{0}                     & 0.4872           & \multicolumn{1}{c|}{0.0116}             & \multicolumn{1}{c|}{\textbf{0.1}}          & 0.5214           \\ \hline
Industrial\_1                     & \multicolumn{1}{c|}{0}                  & \multicolumn{1}{c|}{0}                     & \textbf{0.4832}  & \multicolumn{1}{c|}{0}                  & \multicolumn{1}{c|}{0}                     & 0.4454           \\ \hline
Industrial\_2                     & \multicolumn{1}{c|}{0.0148}             & \multicolumn{1}{c|}{1}                     & \textbf{0.9378}  & \multicolumn{1}{c|}{0.0148}             & \multicolumn{1}{c|}{1}                     & 0.5              \\ \hline
\end{tabular}
}
\caption{A comparative study of \ac{qreg} using standard activation functions versus \ac{pef} activation function on 31 univariate datasets. The best performing algorithm(s) is marked in bold.}
\label{dqr_pef}
\end{table}

\subsubsection{\ac{gan}}
While \ac{gan} with standard activation functions \textit{excelled} in 9 datasets, those with \ac{pef} \textit{performed well} on 3 datasets as seen in Table \ref{gan_pef}. Identical results were obtained for 19 datasets between \ac{gan} with \ac{pef} and \ac{gan} with standard activation functions.
\begin{table}[!htp]
\centering
\resizebox{0.5\textwidth}{!}{
\begin{tabular}{|c|ccc|ccc|}
\hline
\multirow{2}{*}{Dataset} & \multicolumn{3}{c|}{\ac{gan}}                                                                                                    & \multicolumn{3}{c|}{\ac{gan}(\ac{pef})}                                                                                           \\ \cline{2-7} 
                                & \multicolumn{1}{c|}{Precision} & \multicolumn{1}{c|}{Recall} & AUC-ROC & \multicolumn{1}{c|}{Precision} & \multicolumn{1}{c|}{Recall} & AUC-ROC \\ \hline
yahoo1                            & \multicolumn{1}{c|}{0.01}               & \multicolumn{1}{c|}{\textbf{0.5}}            & 0.7              & \multicolumn{1}{c|}{0}                  & \multicolumn{1}{c|}{0}                         & 0.45             \\ \hline
yahoo2                            & \multicolumn{1}{c|}{0}                  & \multicolumn{1}{c|}{0}                       & 0.45             & \multicolumn{1}{c|}{0}                  & \multicolumn{1}{c|}{0}                         & 0.45             \\ \hline
yahoo3                            & \multicolumn{1}{c|}{0}                  & \multicolumn{1}{c|}{0}                       & 0.45             & \multicolumn{1}{c|}{0}                  & \multicolumn{1}{c|}{0}                         & 0.45             \\ \hline
yahoo5                            & \multicolumn{1}{c|}{0}                  & \multicolumn{1}{c|}{0}                       & 0.45             & \multicolumn{1}{c|}{0}                  & \multicolumn{1}{c|}{0}                         & 0.45             \\ \hline
yahoo6                            & \multicolumn{1}{c|}{0}                  & \multicolumn{1}{c|}{\textbf{1}}              & 0.5              & \multicolumn{1}{c|}{0}                  & \multicolumn{1}{c|}{0}                         & 0.45             \\ \hline
yahoo7                            & \multicolumn{1}{c|}{0.01}               & \multicolumn{1}{c|}{\textbf{1}}              & 0.5              & \multicolumn{1}{c|}{0.04}               & \multicolumn{1}{c|}{0.64}                    & 0.77             \\ \hline
yahoo8                            & \multicolumn{1}{c|}{0.01}               & \multicolumn{1}{c|}{\textbf{1}}              & 0.5              & \multicolumn{1}{c|}{0}                  & \multicolumn{1}{c|}{0}                         & 0.45             \\ \hline
yahoo9                            & \multicolumn{1}{c|}{0}                  & \multicolumn{1}{c|}{0}                       & 0.45             & \multicolumn{1}{c|}{0.03}               & \multicolumn{1}{c|}{\textbf{0.63}}           & 0.76             \\ \hline
Speed\_6005                       & \multicolumn{1}{c|}{0}                  & \multicolumn{1}{c|}{1}                       & 0.94             & \multicolumn{1}{c|}{0.00}               & \multicolumn{1}{c|}{1}                       & 0.94             \\ \hline
Speed\_7578                       & \multicolumn{1}{c|}{0.04}               & \multicolumn{1}{c|}{1}                       & 0.95             & \multicolumn{1}{c|}{0.04}               & \multicolumn{1}{c|}{1}                       & 0.95             \\ \hline
Speed\_t4013                      & \multicolumn{1}{c|}{0.01}               & \multicolumn{1}{c|}{\textbf{1}}              & 0.94             & \multicolumn{1}{c|}{0}                  & \multicolumn{1}{c|}{0}                         & 0.45             \\ \hline
TravelTime\_387                   & \multicolumn{1}{c|}{0}                  & \multicolumn{1}{c|}{0}                       & 0.45             & \multicolumn{1}{c|}{0}                  & \multicolumn{1}{c|}{0}                         & 0.45             \\ \hline
TravelTime\_451                   & \multicolumn{1}{c|}{0}                  & \multicolumn{1}{c|}{0}                       & 0.45             & \multicolumn{1}{c|}{0}                  & \multicolumn{1}{c|}{0}                         & 0.45             \\ \hline
Occupancy\_6005                   & \multicolumn{1}{c|}{0}                  & \multicolumn{1}{c|}{0}                       & 0.45             & \multicolumn{1}{c|}{0}                  & \multicolumn{1}{c|}{0}                         & 0.45             \\ \hline
Occupancy\_t4013                  & \multicolumn{1}{c|}{0}                  & \multicolumn{1}{c|}{0}                       & 0.45             & \multicolumn{1}{c|}{0}                  & \multicolumn{1}{c|}{0}                         & 0.45             \\ \hline
yahoo\_syn1                       & \multicolumn{1}{c|}{0}                  & \multicolumn{1}{c|}{0}                       & 0.45             & \multicolumn{1}{c|}{0}                  & \multicolumn{1}{c|}{0}                         & 0.45             \\ \hline
yahoo\_syn2                       & \multicolumn{1}{c|}{0}                  & \multicolumn{1}{c|}{0}                       & 0.45             & \multicolumn{1}{c|}{0}                  & \multicolumn{1}{c|}{0}                         & 0.45             \\ \hline
yahoo\_syn3                       & \multicolumn{1}{c|}{0}                  & \multicolumn{1}{c|}{0}                       & 0.45             & \multicolumn{1}{c|}{0}                  & \multicolumn{1}{c|}{0}                         & 0.45             \\ \hline
yahoo\_syn5                       & \multicolumn{1}{c|}{0.01}               & \multicolumn{1}{c|}{\textbf{1}}              & 0.5              & \multicolumn{1}{c|}{0}                  & \multicolumn{1}{c|}{0}                         & 0.45             \\ \hline
yahoo\_syn6                       & \multicolumn{1}{c|}{0.01}               & \multicolumn{1}{c|}{\textbf{1}}              & 0.5              & \multicolumn{1}{c|}{0}                  & \multicolumn{1}{c|}{0}                         & 0.45             \\ \hline
yahoo\_syn7                       & \multicolumn{1}{c|}{0.01}               & \multicolumn{1}{c|}{\textbf{1}}              & 0.5              & \multicolumn{1}{c|}{0.04}               & \multicolumn{1}{c|}{0.33}                    & 0.62             \\ \hline
yahoo\_syn8                       & \multicolumn{1}{c|}{0.01}               & \multicolumn{1}{c|}{\textbf{1}}              & 0.5              & \multicolumn{1}{c|}{0}                  & \multicolumn{1}{c|}{0}                         & 0.45             \\ \hline
yahoo\_syn9                       & \multicolumn{1}{c|}{0}                  & \multicolumn{1}{c|}{0}                       & 0.45             & \multicolumn{1}{c|}{0.02}               & \multicolumn{1}{c|}{\textbf{0.17}}           & 0.53             \\ \hline
aws1                              & \multicolumn{1}{c|}{0}                  & \multicolumn{1}{c|}{0}                       & 0.45             & \multicolumn{1}{c|}{0}                  & \multicolumn{1}{c|}{0}                         & 0.45             \\ \hline
aws2                              & \multicolumn{1}{c|}{0}                  & \multicolumn{1}{c|}{0}                       & 0.41             & \multicolumn{1}{c|}{0}                  & \multicolumn{1}{c|}{0}                         & 0.41             \\ \hline
aws3                              & \multicolumn{1}{c|}{0}                  & \multicolumn{1}{c|}{0}                       & 0.45             & \multicolumn{1}{c|}{0}                  & \multicolumn{1}{c|}{0}                         & 0.45             \\ \hline
aws\_syn1                         & \multicolumn{1}{c|}{0}                  & \multicolumn{1}{c|}{0}                       & 0.45             & \multicolumn{1}{c|}{0}                  & \multicolumn{1}{c|}{0}                         & 0.45             \\ \hline
aws\_syn2                         & \multicolumn{1}{c|}{0}                  & \multicolumn{1}{c|}{0}                       & 0.41             & \multicolumn{1}{c|}{0}                  & \multicolumn{1}{c|}{0}                         & 0.41             \\ \hline
aws\_syn3                         & \multicolumn{1}{c|}{0}                  & \multicolumn{1}{c|}{0}                       & 0.45             & \multicolumn{1}{c|}{0}                  & \multicolumn{1}{c|}{0}                         & 0.45             \\ \hline
Industrial\_1                     & \multicolumn{1}{c|}{0}                  & \multicolumn{1}{c|}{0}                       & 0.41             & \multicolumn{1}{c|}{0}                  & \multicolumn{1}{c|}{0}                         & 0.41             \\ \hline
Industrial\_2                     & \multicolumn{1}{c|}{0}                  & \multicolumn{1}{c|}{0}                       & 0.41             & \multicolumn{1}{c|}{0.13}               & \multicolumn{1}{c|}{\textbf{1}}              & 0.95             \\ \hline
\end{tabular}
}
\caption{A comparative study of \ac{gan} using standard activation functions versus \ac{pef} activation function on 31 univariate datasets. The best performing algorithm(s) is marked in bold.}
\label{gan_pef}
\end{table}
\subsubsection{\ac{autoencoder}}
The evaluation of \ac{autoencoder}, revealed identical results for models employing standard activation functions and those utilizing \ac{pef} activation. Using \ac{pef} did not alter the performance of \ac{autoencoder} based on the considered metrics.

\section{Conclusion}
\label{sec:sec4}
The study undertakes the crucial task of benchmarking and rigorous evaluation of a diverse set of anomaly detection algorithms to address the pressing need for robust anomaly detection in complex mission-critical systems. The identification of anomalies is pivotal for ensuring the resilience of such systems, and our study specifically tackled the challenge of imbalanced class distribution in operational data, where anomalies are infrequent but critical events nonetheless.

Apart from the comprehensive benchmark study covering 104 datasets, the unbiased evaluation provides a fair comparison across classical \ac{ml}, \ac{dl}, and outlier detection methods, the study offers useful insights and raises important questions. 
The results demonstrate that DL algorithms tend to dominate in multivariate datasets, primarily due to their data-hungry nature. The larger size of multivariate datasets provided the necessary data for DL models to perform comparatively well. However, it is important to ponder over the efficacy of the recent SOTA \ac{dl} methods such as \ac{ftt}, \ac{prenet}, \ac{deepsad} etc. which require anomaly contamination in the training set. Anomalies are deviations from the normal patterns in data and are not known a priori. In addition, the comparatively longer training time and greater resource consumption raise questions about the utility and adaptability of the SOTA \ac{dl} methods in realistic settings.
For instance, \ac{ftt}, one of the best-performing DL algorithms, excelled across many metrics in multivariate settings. However, in univariate datasets, where the data size is relatively smaller, unsupervised tree-based methods like \ac{mgbtai}, which are less dependent of the availability of data, outperformed DL models. Out of the 31 univariate datasets, \ac{mgbtai} achieved the highest precision in 25 datasets, the highest recall in 13 datasets, the highest F1 score in 12 datasets and the highest AUC-ROC score in 11 datasets, demonstrating its strong performance in these settings.
The findings underscore the importance of unsupervised algorithms such as \ac{mgbtai} in real-world applications. Given the challenges in acquiring well-labeled data, especially in anomaly detection tasks where anomalies are rare, unsupervised methods like \ac{mgbtai} offer a critical advantage by eliminating the need for labeled datasets while maintaining the ability to effectively detect anomalies.
In addition to performance differences, DL algorithms also carry the baggage of significant computational costs and substantial computational resources.
This trade-off between performance and computational efficiency is a crucial factor when selecting an anomaly detection method, especially in time-sensitive and resource-constrained environments.
Tree-based methods use clustering to group similar data and separate out small-sized clusters as anomalies and iterate this over multiple generations. Tree based algorithms conjecture that anomalies are susceptible to isolation in a minimum number of partitioning compared to normal datasets. This, no requirement of training and no assumption on the number of anomalies, type of data/distribution gives them an advantage over DNNs in realistic settings. However, what we wanted to drive home throughout the paper is that, despite the dominance of DL based methods in all branches of AI, a lesser known method such as the tree based one has performed well and more importantly, is better positioned for adoption in real-time systems. We believe this is an important point to ponder for the community as the findings have been a revelation to us.
\par Real world applicability of anomaly detection system remains a challenge because of the complexities in Recall-Precision trade-off. It is therefore, pertinent to discuss some use-case scenarios of tree based algorithms (for instance \ac{mgbtai} and \ac{dbtai}) based on their observed performance in this study. Between the two, \ac{dbtai} achieves better recall, often the SOTA performance and comparable or slightly worse performance in Precision when compared to the recent SOTA \ac{dl} methods. \ac{mgbtai} on the other hand, achieves a better balance between Precision and Recall. Thus, \ac{dbtai} might be used in safety-critical systems (Industrial Systems, PET Scanners etc.). High Recall, high Precision detection is desirable but is a myth. Non-safety critical systems such as detecting spam emails is an example of High Precision and low Recall detection. Other examples of such a system include recommendation of profitable stocks, detection of critical bugs in software etc.
\par Looking ahead, this benchmark study lays a robust groundwork for future advancements in the realm of anomaly detection for robust anomaly detection in complex mission-critical systems. Further exploration of hybrid models, leveraging the strengths of various algorithms, could enhance adaptability to evolving challenges in complex systems. Explainable anomaly detection in mission-critical systems can also be interesting future work.
\bibliography{references,modified}

\bibliographystyle{IEEEtran}

\end{document}